\documentclass{llncs}
\usepackage{verbatim}
\usepackage{paralist}
\usepackage[utf8]{inputenc}
\usepackage{fixltx2e} 
\usepackage{enumitem}
\usepackage{multirow}

\usepackage{hyperref}
\hypersetup{
    bookmarks=true,         
    unicode=false,          
    pdftoolbar=true,        
    pdfmenubar=true,        
    pdffitwindow=false,     
    pdfstartview={FitH},    
    pdftitle={},            
    pdfauthor={},           
    pdfnewwindow=true,      
    colorlinks=true,        
    linkcolor=black,        
    citecolor=black,        
    filecolor=black,        
    urlcolor=black          
}

\usepackage{listings} 
\begin{document}

\pagestyle{headings}  

\mainmatter              

\title{Controlled Natural Language Generation from a Multilingual FrameNet-based Grammar} 

\titlerunning{CNL Generation from a Multilingual FrameNet-based Grammar}

\author{Dana Dann\'{e}lls \and Normunds Gruzitis}

\authorrunning{Dann\'{e}lls and Gruzitis} 

\institute{
Spr\r{a}kbanken, University of Gothenburg\\
Department of Computer Science and Engineering, University of Gothenburg\\
\email{dana.dannells@svenska.gu.se, normunds.gruzitis@cse.gu.se}
}

\maketitle

\begin{abstract} 
This paper presents a currently bilingual but potentially multilingual FrameNet-based grammar library implemented in Grammatical Framework. 
The contribution of this paper is two-fold. First, it offers a methodological approach to automatically generate the grammar based on semantico-syntactic valence patterns extracted from FrameNet-annotated corpora. Second, it provides a proof of concept for two use cases illustrating how the acquired multilingual grammar can be exploited in different CNL applications in the domains of arts and tourism. 

\end{abstract}

\section{Introduction}

Two years ago, at CNL~2012, a conception of a general-purpose semantic grammar based on FrameNet (FN) was proposed~\cite{GruzitisEtAl2012} to facilitate the development of multilingual controlled natural language (CNL) applications in Grammatical Framework (GF). GF~\cite{Ranta2004}, a type-theoretical grammar formalism and a toolkit, provides a wide-coverage resource grammar library (RGL) for nearly 30 languages that implement a shared syntactic API~\cite{Ranta2009}. The idea behind the FN-based grammar is to provide a frame semantic abstraction layer, a shared semantic API, over the syntactic RGL.

Following this proposal, a shared abstract syntax of wide-coverage English and Swedish semantic grammars has been recently extracted from FN-annotated corpora~\cite{DannellsAndGruzitis2014}. 
In this work, we take this approach one step further, and the contribution of this paper is two-fold. First, we offer a methodological approach to automatically generate concrete syntaxes based on the extracted abstract syntax. Second, we provide a proof of concept for two use cases illustrating how the acquired multilingual grammar can be exploited in different CNL applications in the domains of arts and tourism. Although we focus on English and Swedish, the same approach is intended to be applicable to other languages as well.

The future potential of our work is to provide a means for multilingual verbalization of FN-annotated databases that have been populated in information extraction processes by FN-based semantic parsers and that potentially can be mapped with the FN-based API automatically~\cite{Barzdins2014}.

\section{Background}
\label{sec:back}

\subsection{FrameNet (FN)}
\label{ssec:fn}

FrameNet is a lexico-semantic resource based on the theory of frame semantics~\cite{FillmoreEtAl2003}. According to this theory, a semantic \emph{frame} representing a cognitive scenario is characterized in terms of \emph{frame elements}~(FE) and is evoked by target words called \emph{lexical units}~(LU). 
An LU entry carries semantic and syntactic valence information about the possible realizations of FEs. The syntactic and semantic valence patterns are derived from FN-annotated corpora. FEs are classified in \emph{core} and \emph{non-core} FEs. A set of core FEs uniquely characterize the frame and syntactically correspond to verb arguments, in contrast to non-core FEs (adjuncts) which can be instantiated in many other frames. In this paper, we consider only those frames for which there is at least one corpus example where the frame is evoked by a verb. The frame-based grammar currently covers only core FEs. 

The FrameNet approach provides a benchmark for representing large amounts of word senses and word usage patterns through the linguistic annotation of corpus examples, therefore the exploitation of FN-like resources has been appealing for a range of advanced NLP applications such as semantic parsing~\cite{DasEtAl2013}, information extraction~\cite{MoschittiEtAl2003} and natural language generation~\cite{RothAndFrank2009}. There are available computationally oriented FNs for German, Japanese, Spanish~\cite{Boas2009} and Swedish~\cite{BorinEtAl2010}. More initiatives exist for other languages. In this paper, we consider two FNs: the original Berkeley FrameNet (BFN)~\cite{FillmoreEtAl2003} and the Swedish FrameNet (SweFN)~\cite{BorinEtAl2010}.

BFN version 1.5 defines more than 1,000 frames,\footnote{\url{https://framenet.icsi.berkeley.edu/}} of which 556 are evoked by around 3,200 verb LUs in more than 68,500 annotated sentences~\cite{DannellsAndGruzitis2014}. Although BFN has been developed for English, its inventory of frames and FEs is being reused for many other FNs~\cite{Boas2009}. Hence, the abstract semantic layer of BFN can be seen as an interlingua for linking different FNs.

SweFN mostly uses the BFN frame inventory, however, around 50 additional frames have been introduced in SweFN, and around 15 BFN frames have been modified (in terms of FEs). The SweFN development version contains more than 900 frames of which 638 are evoked by around 2,300 verb LUs in more than 3,700 annotated sentences~\cite{DannellsAndGruzitis2014}.\footnote{\url{http://spraakbanken.gu.se/swefn/} (a snapshot taken in February 2014)}

\subsection{Grammatical Framework (GF)}
\label{ssec:gf}

The presented grammar is implemented in GF, a categorial grammar formalism specialized for multilingual (parallel) grammars~\cite{Ranta2004}. One of the key features of GF grammars is the separation between an abstract syntax and concrete syntaxes. The abstract syntax defines the language-independent structure, the semantics of a domain-specific application grammar or a general-purpose grammar library, while the concrete syntaxes define the language-specific syntactic and lexical realization of the abstract syntax.

Remarkably, GF is not only a grammar formalism or programming language. It also provides a general-purpose resource grammar library (RGL) for nearly 30 languages that implement the same abstract syntax, a shared syntactic API~\cite{Ranta2009}. The use of the shared syntactic types and functions allows for rapid and rather flexible development of multilingual application grammars without the need of specifying low-level details like inflectional paradigms and syntactic agreement.

\section{FrameNet-based Grammar}
\label{sec:fn-api}

The language-independent conceptual layer of FrameNet, i.e. frames and FEs, is defined in the abstract syntax of the multilingual FN-based grammar, while the language-specific lexical layers, i.e. the surface realization of frames and LUs, are defined in concrete syntaxes.\footnote{\url{http://www.grammaticalframework.org/framenet/}} 
The syntactic API of RGL is used for generalizing and unifying the syntactic types and constructions used in different FNs, which facilitates porting the implementation to other languages. The FN-based grammar, in turn, provides a frame semantic abstraction layer to RGL, so that the application grammar developer can primarily manipulate with plain semantic constructors in combination with some simple syntactic constructors instead of comparatively complex syntactic constructors for building verb phrases (VP). Moreover, the frame constructors can be typically specified for all languages at once in the shared concrete syntax (functor) of an application grammar.

\subsection{Abstract Syntax}
\label{ssec:abstract}

Following a recently proposed approach~\cite{DannellsAndGruzitis2014}, we have extracted a set of shared semantico-syntactic frame valence patterns from the annotated sentences in BFN and SweFN. For instance, the shared valence patterns for the frame $\mathsf{Desiring}$ are:

\begingroup
\fontsize{9pt}{11pt}\selectfont
\begin{enumerate}
\item[] $\mathsf{Desiring/V_{Act}}$ $\mathsf{Experiencer/NP_{Subj}}$ $\mathsf{Focal\_participant/Adv}$
\item[] $\mathsf{Desiring/V2_{Act}}$ $\mathsf{Experiencer/NP_{Subj}}$ $\mathsf{Focal\_participant/NP_{DObj}}$
\item[] $\mathsf{Desiring/VV_{Act}}$ $\mathsf{Event/VP}$ $\mathsf{Experiencer/NP_{Subj}}$
\end{enumerate}
\endgroup

\noindent which correspond, for instance, to these annotated examples in BFN:\footnote{The actual BFN phrase types are generalized by RGL types.}

\begingroup
\fontsize{9pt}{11pt}\selectfont
\begin{enumerate}
\item[] {[\emph{Dexter}]}\textsubscript{$\mathsf{Experiencer/NP}$} [\emph{YEARNED}]\textsubscript{$\mathsf{V}$} [\emph{for a cigarette}]\textsubscript{$\mathsf{Focal\_participant/Adv}$}
\item[] {[\emph{she}]}\textsubscript{$\mathsf{Experiencer/NP}$} [\emph{WANTS}]\textsubscript{$\mathsf{V2}$} [\emph{a protector}]\textsubscript{$\mathsf{Focal\_participant/NP}$}
\item[] {[\emph{I}]}\textsubscript{$\mathsf{Experiencer/NP}$} \emph{would n't} [\emph{WANT}]\textsubscript{$\mathsf{VV}$} [\emph{to know}]\textsubscript{$\mathsf{Event/VP}$}
\end{enumerate}
\endgroup

In contrast to the previous experiment~\cite{DannellsAndGruzitis2014}, where the focus was on the abstract grammar, here 
we generate the concrete syntaxes taking the syntactic roles for FEs of type $\mathsf{NP}$ into account: subject ($\mathsf{Subj}$), direct object ($\mathsf{DObj}$) and indirect object ($\mathsf{IObj}$). Thus, we also consider the grammatical voice ($\mathsf{Act}$/$\mathsf{Pass}$) in the pattern comparison, as well as the target verb type deduced from the syntactic types and roles of involved FEs. Additionally, we handle FEs of common types of subclauses (generalized to $\mathsf{S}$, embedded sentences), as well as finite and gerundive VPs, and PPs where the preposition governs a wh-clause or a gerundive VP, so that the fraction of skipped BFN examples is reduced form 14\% to 4\%, and no SweFN examples are skipped.

The extracted sets of valence patterns usually vary across languages depending on corpora. For multilingual applications we are primarily interested in valence patterns whose implementation can be generated for all considered languages. Thus, we focus on valence patterns that are shared between FNs. The multilingual criteria also help in reducing the number of incorrect patterns due to annotation errors introduced by the automatic POS tagging and syntactic parsing. However, patterns that are not verified across FNs could be separated into FN-specific extra modules of the grammar.

To find a representative yet condensed set of shared valence patterns, we compare the extracted patterns by subsumption instead of exact match~\cite{DannellsAndGruzitis2014}. Pattern $\mathsf{A}$ subsumes pattern $\mathsf{B}$ if $\mathsf{A.frame = B.frame}$, $\mathsf{A.verbType = B.verbType}$, $\mathsf{A.voice = B.voice}$, and $\mathsf{B.FEs \subseteq A.FEs}$ (taking into account the syntactic types and roles). If a pattern of $\mathsf{FN_1}$ is subsumed by a pattern of $\mathsf{FN_2}$, it is added to the shared set (and vice versa). In the final set, patterns which are subsumed by other shared patterns are removed. To reduce the propagation of annotation errors even more, we filter out once used BFN valence patterns before performing the cross-FN pattern comparison.\footnote{A similar pre-filtering is currently not reasonable for SweFN due to its small size.}

In the result, from around 66,800 annotated sentences in BFN and around 4,100 annotated sentences in SweFN, we have extracted a set of 717 shared semantico-syntactic valence patterns covering 423 frames.

Frame valence patterns are declared in the grammar as functions (henceforth called frame functions) that take one or more core FEs and one verb as arguments. For each frame, the set of core FEs is often split into several alternative functions according to the corpus evidence.\footnote{It is often unlikely that all core FEs can be used in the same sentence.} Different subsets of core FEs may require different types of target verbs. We also differentiate between functions that return clauses in the passive voice from functions that return active voice clauses because the subject and object FEs swap their syntactic roles and/or the order (which otherwise is not reflected in the abstract syntax). If the verb type and voice suffixes are not sufficient to make the function name unique, a discriminative number is added as well. For instance, consider the following abstract functions derived from the above given valence patterns:\footnote{Note that $\mathsf{Desiring\_V2\_Pass}$ is not directly acquired from a shared pattern; missing passive voice patterns could be derived from the corresponding active voice patterns. Also note that the syntactic roles are not reflected in the abstract syntax; they are used to generate the implementation of frame functions in the concrete syntaxes.}

\begingroup
\fontsize{9pt}{11pt}\selectfont
\begin{enumerate}
\item[] $\mathit{fun}$ $\mathsf{Desiring\_V : Experiencer\_NP \to Focal\_participant\_Adv \to V \to Clause}$
\item[] $\mathit{fun}$ $\mathsf{Desiring\_V2\_Act : Experiencer\_NP \to Focal\_participant\_NP \to V2 \to Clause}$
\item[] $\mathit{fun}$ $\mathsf{Desiring\_V2\_Pass : Experiencer\_NP \to Focal\_participant\_NP \to V2 \to Clause}$
\item[] $\mathit{fun}$ $\mathsf{Desiring\_VV : Event\_VP \to Experiencer\_NP \to VV \to Clause}$
\end{enumerate}
\endgroup

In GF, constituents and features of phrases are stored in objects of record types, and functions are applied to such objects to construct phrase trees. In the abstract syntax, both argument types and the value type of a function are separated by right associative arrows, i.e. all functions are curried. Arguments of a frame function are combined into an object of type $\mathsf{Clause}$ that differs form the RGL type $\mathsf{Cl}$. A $\mathsf{Clause}$ whose linearization type is $\mathsf{\{np : NP ; vp : VP\}}$ comprises two constituents of RGL types. It is a deconstructed $\mathsf{Cl}$ where the subject NP is separated from the rest of the clause. The motivation for this is to allow for nested frames (see Section~\ref{ssec:phrasebook}) and for adding non-core FEs before combining the NP and VP parts into a clause (see Section~\ref{ssec:museum}).

In the FN-based grammar, FEs are declared as semantic categories (types) that are subcategorized by RGL types, and these discriminators are also encoded by suffixes in FE names to keep the names unique, for instance:

\begingroup
\fontsize{9pt}{11pt}\selectfont
\begin{enumerate}
\item[] $\mathit{cat}$ $\mathsf{Experiencer\_NP}$
\end{enumerate}
\endgroup

Note that the FE $\mathsf{Focal\_participant}$ is typically realized as a noun phrase (NP), but some intransitive verbs require it as a prepositional phrase (PP), hence this FE is subcategorized using the RGL types $\mathsf{NP}$ and $\mathsf{Adv}$ (adverbial modifier). In GF, the type $\mathsf{Adv}$ covers both adverbs and PPs, and there is no separate type for PPs. Also note that the word order is not specified in the abstract syntax (FEs in the function type signatures are given alphabetically), and all FE arguments are specified in concrete syntaxes as optional, i.e. any FE can be an empty phrase if it is not expressed in the sentence.

The frame-evoking target verb, either intransitive ($\mathsf{V}$), transitive ($\mathsf{V2}$) or ditransitive ($\mathsf{V3}$), is always given as the last, mandatory argument. We additionally differentiate two special cases of transitive verbs: verb-phrase-complement verbs ($\mathsf{VV}$) and sentence-complement verbs ($\mathsf{VS}$), as well as a special case for each of them allowing also for an indirect object ($\mathsf{V2V}$ and $\mathsf{V2S}$ respectively).

LUs are represented as functions that take no arguments. To distinguish between different senses and types of LUs, the verb type and the frame name is added to lexical function names, for instance:

\begingroup
\fontsize{9pt}{11pt}\selectfont
\begin{enumerate}
\item[] (Eng) $\mathit{fun}$ $\mathsf{want\_VV\_Desiring : VV}$
\item[] (Swe) $\mathit{fun}$ $\mathsf{vilja\_VV\_Desiring : VV}$
\end{enumerate}
\endgroup

However, LUs between BFN and SweFN are not directly aligned, therefore an FN-specific lexicon is generated for each language containing more than 3,300 entries for English and more than 1,100 entries for Swedish. The domain-specific translation equivalents can be aligned in application grammars.

We assume that verbs of the same type evoking the same frame share a set of generalized syntactic valence patterns. Patterns requiring, for instance, a transitive verb cannot be evoked by an intransitive verb. Otherwise, the current approach does not limit the set of verbs that can evoke a frame, and the set of prepositions that can be used for an FE if it is realized as a PP. We expect that appropriate verbs and prepositions are specified by the application grammar that uses the FN-based grammar as an API. Hence, this approach allows to evoke a frame by a metaphor, i.e. an LU that normally evokes another frame.

\subsection{Concrete Syntaxes}
\label{ssec:concrete}

The exact behaviour of the types and functions declared in the abstract syntax is defined in the concrete syntax for each language.

The mapping from the semantic FN types to the syntactic RGL types is straightforward and is shared for all languages in a functor, for instance:

\begingroup
\fontsize{9pt}{11pt}\selectfont
\begin{enumerate}
\item[] $\mathit{lincat}$ $\mathsf{Focal\_participant\_NP = Maybe}$ $\mathsf{NP}$
\item[] $\mathit{lincat}$ $\mathsf{Focal\_participant\_Adv = Maybe}$ $\mathsf{Adv}$
\end{enumerate}
\endgroup

To allow for optional FEs (verb arguments that might not be expressed in the sentence), all linearization types are of type $\mathsf{Maybe}$ whose behaviour is similar to the analogous type in Haskell: a value of type $\mathsf{Maybe}$ $\mathit{x}$ either contains a value of type $\mathit{x}$ (represented as $\mathsf{Just}$ $\mathit{x}$), or it is empty (represented as $\mathsf{Nothing}$).

To implement the frame functions, particularly to fill the VP part of $\mathsf{Clause}$ objects, RGL constructors are applied to the arguments depending on their RGL types and syntactic roles. The implementation of functions declared in the previous section is systematically generated for English and Swedish as follows:

\begingroup
\fontsize{9pt}{11pt}\selectfont
\begin{enumerate}
\item[] $\mathit{lin}$ $\mathsf{Desiring\_V}$ $\mathrm{experiencer}$ $\mathrm{focal\_participant}$ $\mathrm{v = \{}$
\begin{enumerate}
\item[] $\mathrm{np =}$ $\mathsf{fromMaybe}$ $\mathsf{NP}$ $\mathrm{experiencer}$ $\mathrm{;}$
\item[] $\mathrm{vp =}$ $\mathsf{mkVP}$ $\mathsf{(mkVP}$ $\mathrm{v}$$\mathsf{)}$ $\mathsf{(fromMaybe}$ $\mathsf{Adv}$ $\mathrm{focal\_participant}$$\mathsf{)}$ $\mathrm{\}}$
\end{enumerate}
\end{enumerate}
\endgroup

\begingroup
\fontsize{9pt}{11pt}\selectfont
\begin{enumerate}
\item[] $\mathit{lin}$ $\mathsf{Desiring\_V2\_Act}$ $\mathrm{experiencer}$ $\mathrm{focal\_participant}$ $\mathrm{v2 = \{}$
\begin{enumerate}
\item[] $\mathrm{np =}$ $\mathsf{fromMaybe}$ $\mathsf{NP}$ $\mathsf{experiencer}$ $\mathrm{;}$
\item[] $\mathrm{vp =}$ $\mathsf{mkVP}$ $\mathrm{v2}$ $\mathsf{(fromMaybe}$ $\mathsf{NP}$ $\mathrm{focal\_participant}$$\mathsf{)}$ $\mathrm{\}}$
\end{enumerate}
\end{enumerate}
\endgroup

\begingroup
\fontsize{9pt}{11pt}\selectfont
\begin{enumerate}
\item[] $\mathit{lin}$ $\mathsf{Desiring\_V2\_Pass}$ $\mathrm{experiencer}$ $\mathrm{focal\_participant}$ $\mathrm{v2 = \{}$
\begin{enumerate}
\item[] $\mathrm{np =}$ $\mathsf{fromMaybe}$ $\mathsf{NP}$ $\mathrm{focal\_participant}$ $\mathrm{;}$
\item[] $\mathrm{vp =}$ $\mathsf{mkVP}$ $\mathsf{(passiveVP}$ $\mathrm{v2}$$\mathsf{)}$ $\mathsf{(mkAdv}$ $\mathsf{by8agent\_Prep}$ $\mathsf{(fromMaybe}$ $\mathsf{NP}$ $\mathrm{experiencer}$$\mathsf{))}$
\end{enumerate}
\item[] $\mathrm{\}}$
\end{enumerate}
\endgroup

\begingroup
\fontsize{9pt}{11pt}\selectfont
\begin{enumerate}
\item[] $\mathit{lin}$ $\mathsf{Desiring\_VV}$ $\mathrm{event}$ $\mathrm{experiencer}$ $\mathrm{vv = \{}$
\begin{enumerate}
\item[] $\mathrm{np =}$ $\mathsf{fromMaybe}$ $\mathsf{NP}$ $\mathrm{experiencer}$ $\mathrm{;}$
\item[] $\mathrm{vp =}$ $\mathsf{mkVP}$ $\mathsf{(mkVV}$ $\mathrm{vv}$$\mathsf{)}$ $\mathsf{(fromMaybe}$ $\mathsf{VP}$ $\mathrm{event}$$\mathsf{)}$ $\mathrm{\}}$
\end{enumerate}
\end{enumerate}
\endgroup

Apart from RGL constructors ($\mathsf{mkVP}$, $\mathsf{mkVV}$, $\mathsf{passiveVP}$, $\mathsf{mkAdv}$, etc.\footnote{\url{http://www.grammaticalframework.org/lib/doc/synopsis.html}}), a helper function $\mathsf{fromMaybe}$ is used to handle the potentially optional FEs. This function takes a $\mathsf{Maybe}$ value and returns an empty phrase of the specified type if the $\mathsf{Maybe}$ value is empty ($\mathsf{Nothing}$); otherwise it returns the $\mathsf{Maybe}$ value.

The RGL-based code templates used to implement the above functions can be reused for many other frame functions. Given the 717 extracted shared semantico-syntactic valence patterns, there are only 25 syntactic valence patterns that match all 717 patterns if we consider only the syntactic types and roles of FEs, and the grammatical voice the roles depend on. These patterns (except 5 once used) are listed in Table~\ref{tab:freq} that shows that the syntactic patterns underlying functions $\mathsf{Desiring\_V}$, $\mathsf{Desiring\_V2\_Act}$, $\mathsf{Desiring\_V2\_Pass}$ and $\mathsf{Desiring\_VV}$ already cover 55\% of all shared patterns. For the same verb types, similar syntactic patterns (RGL-based code templates) cover another 39\% of frame functions. The similar templates can be derived in several (incl. combined) ways:

\begin{itemize}[noitemsep]
\item more adverbial modifiers can be added by recursive calls of the respective $\mathsf{mkVP}$ constructor, or modifiers can be removed at all;
\item the NP part of the return values can be fixed to an empty NP if no FE is expected to fill the subject role (e.g. due to examples in the imperative mood; however, a missing subject FE could be often automatically added);
\item in the passive voice, the direct object can be possibly fixed to an empty NP.
\end{itemize}

\begin{table}[h]
\tabcolsep 1.5pt
\begin{center}
\begin{tabular}{|lllr||lllr|}
\hline
Verb & Voice & FE types and roles & Freq. & Verb & Voice & FE types and roles & Freq. \\
\hline
V2 & Act & NP\textsubscript{DObj} NP\textsubscript{Subj} & 238 & V & Act & Adv & 8 \\
V & Act & Adv NP\textsubscript{Subj} & 138 & V2 & Act & Adv NP\textsubscript{DObj} & 8 \\
V2 & Pass & NP\textsubscript{Subj} & 70 & V2V & Act & NP\textsubscript{IObj} NP\textsubscript{Subj} VP & 5 \\
V & Act & NP\textsubscript{Subj} & 65 & VS & Pass & S & 3 \\
V2 & Act & Adv NP\textsubscript{DObj} NP\textsubscript{Subj} & 62 & V & Act & Adv Adv Adv NP\textsubscript{Subj} & 2 \\
V2 & Pass & Adv NP\textsubscript{Subj} & 31 & V2 & Act & Adv Adv NP\textsubscript{DObj} NP\textsubscript{Subj} & 2 \\
VS & Act & NP\textsubscript{Subj} S & 26 & V2 & Pass & Adv & 2 \\
VV & Act & NP\textsubscript{Subj} VP & 18 & V2 & Pass & Adv Adv NP\textsubscript{Subj} & 2 \\
V & Act & Adv Adv NP\textsubscript{Subj} & 14 & V3 & Act & NP\textsubscript{IObj} NP\textsubscript{Subj} & 2 \\
V2 & Act & NP\textsubscript{DObj} & 14 & VS & Act & Adv NP\textsubscript{Subj} S & 2 \\
\hline
\end{tabular}
\end{center}
\caption{\label{tab:freq} Syntactic valence patterns matching the shared semantico-syntactic patterns}
\end{table}

The remaining 6\% of the shared patterns represent the use of other verb types: $\mathsf{V3}$, $\mathsf{V2V}$, $\mathsf{VS}$ and $\mathsf{V2S}$. Basic code templates that are reused to implement the corresponding frame functions (VP parts) are illustrated by these examples:

\begingroup
\fontsize{9pt}{11pt}\selectfont
\begin{enumerate}
\item[] $\mathsf{mkVP}$ $\mathrm{v3}$ $\mathsf{(fromMaybe}$ $\mathsf{NP}$ $\mathrm{recipient}$$\mathsf{)}$ $\mathsf{(fromMaybe}$ $\mathsf{NP}$ $\mathrm{theme}$$\mathsf{)}$
\item[] {\tt -{}-} $\mathsf{Giving}$: [\emph{she}]\textsubscript{$\mathsf{Donor/NP}$} [\emph{handed}]\textsubscript{$\mathsf{V3}$} [\emph{him}]\textsubscript{$\mathsf{Recipient/NP}$} [\emph{the ring}]\textsubscript{$\mathsf{Theme/NP}$}
\end{enumerate}
\endgroup

\begingroup
\fontsize{9pt}{11pt}\selectfont
\begin{enumerate}
\item[] $\mathsf{mkVP}$ $\mathrm{vs}$ $\mathsf{(fromMaybe}$ $\mathsf{S}$ $\mathrm{message}$$\mathsf{)}$
\item[] {\tt -{}-} $\mathsf{Hear}$: [\emph{we}]\textsubscript{$\mathsf{Hearer/NP}$} [\emph{heard}]\textsubscript{$\mathsf{VS}$} [\emph{it was a good school}]\textsubscript{$\mathsf{Message/S}$}
\end{enumerate}
\endgroup

\begingroup
\fontsize{9pt}{11pt}\selectfont
\begin{enumerate}
\item[] $\mathsf{mkVP}$ $\mathrm{v2v}$ $\mathsf{(fromMaybe}$ $\mathsf{NP}$ $\mathrm{addressee}$$\mathsf{)}$ $\mathsf{(fromMaybe}$ $\mathsf{VP}$ $\mathrm{message}$$\mathsf{)}$
\item[] {\tt -{}-} $\mathsf{Request}$: [\emph{UK}]\textsubscript{$\mathsf{Speaker/NP}$} [\emph{urges}]\textsubscript{$\mathsf{V2V}$} [\emph{Savimbi}]\textsubscript{$\mathsf{Addressee/NP}$} [\emph{to keep the peace}]\textsubscript{$\mathsf{Message/VP}$}
\end{enumerate}
\endgroup

\begingroup
\fontsize{9pt}{11pt}\selectfont
\begin{enumerate}
\item[] $\mathsf{mkVP}$ $\mathrm{v2s}$ $\mathsf{(fromMaybe}$ $\mathsf{NP}$ $\mathrm{addressee}$$\mathsf{)}$ $\mathsf{(fromMaybe}$ $\mathsf{S}$ $\mathrm{content}$$\mathsf{)}$
\item[] {\tt -{}-} $\mathsf{Suasion}$: [\emph{he}]\textsubscript{$\mathsf{Speaker/NP}$} [\emph{persuaded}]\textsubscript{$\mathsf{V2S}$} [\emph{himself}]\textsubscript{$\mathsf{Addressee/NP}$} [\emph{that they helped}]\textsubscript{$\mathsf{Content/S}$}
\end{enumerate}
\endgroup

Note that the RGL type $\mathsf{S}$, embedded declarative sentence, is used only if the subclause can be verbalized using the subjunction \emph{that}; otherwise such FEs are subcategorized as $\mathsf{Adv}$, and the application grammar developer has to specify the subjunction by applying the RGL constructor $\mathsf{mkAdv : Subj \to S \to Adv}$. Also note that FEs of type $\mathsf{VP}$ or $\mathsf{S}$, or $\mathsf{Adv}$ encapsulating an $\mathsf{S}$ represent nested frames. We use the type $\mathsf{S}$ instead of $\mathsf{Cl}$ to allow for specifying sentence level parameters like tense, anteriority and polarity of the nested frames.

The implementation of frame functions, although currently kept separate for each language, mostly could be shared due to the syntactic abstraction provided by RGL. In general, however, the order of $\mathsf{Adv}$ FEs can differ across languages.

\section{Case Studies}
\label{sec:cases}

We illustrate the use of the FrameNet-based API to GF RGL by re-engineering two existing multilingual GF application grammars: one for translating standard tourist phrases \cite{RantaEtAl2012} and another for generating descriptions of paintings \cite{DannellsEtAl2012}, both developed in the MOLTO project.\footnote{\url{http://www.molto-project.eu/}} In both cases, we preserve the original functionality, and we do not make any changes in the application abstract syntax. Changes affect only the concrete syntaxes of English and Swedish.

\subsection{Phrasebook}
\label{ssec:phrasebook}

Although the Phrasebook grammar covers many idiomatic expressions that cannot be translated using the same frame or for which the FN-based approach would not be suitable at all, it includes around 20 complex clause-building functions that can be handled by the FN-based grammar. To illustrate the use of the semantic API, we re-implement the following Phrasebook functions:

\begingroup
\fontsize{9pt}{11pt}\selectfont
\begin{verbatim}
ALive   : Person -> Country -> Action  -- e.g. `we live in Sweden'
AWant   : Person -> Object -> Action   -- e.g. `I want a pizza'
AWantGo : Person -> Place -> Action    -- e.g. `I want to go to a museum'
\end{verbatim}
\endgroup

\noindent by applying the frame functions $\mathsf{Desiring\_V2\_Act}$ and $\mathsf{Desiring\_VV}$ introduced in Section \ref{sec:fn-api}, and some additional functions:

\begingroup
\fontsize{9pt}{11pt}\selectfont
\begin{verbatim}
Motion_V_2    : Goal_Adv -> Source_Adv -> Theme_NP -> Clause
Possession_V2 : Owner_NP -> Possession_NP -> Clause
Residence_V   : Location_Adv -> Resident_NP -> Clause
\end{verbatim}
\endgroup

By using RGL constructors, $\mathsf{ALive}$ is implemented for English, Swedish and other languages in the same way, except that different verbs are used:

\begingroup
\fontsize{9pt}{11pt}\selectfont
\begin{verbatim}
ALive p co = mkCl p.name (mkVP (mkVP (mkV "live")) (mkAdv in_Prep co))
ALive p co = mkCl p.name (mkVP (mkVP (mkV "bo")) (mkAdv in_Prep co))
\end{verbatim}
\endgroup

First, the language-specific verbs can be factored out by introducing a shared abstract verb in the domain lexicon (e.g. $\mathsf{live\_V}$ that links $\mathsf{live\_V\_Residence}$ and $\mathsf{bo\_V\_Residence}$). Second, the implementation of $\mathsf{ALive}$ can be done in a shared functor by using the FN-based API:

\begingroup
\fontsize{9pt}{11pt}\selectfont
\begin{verbatim}
ALive p co = let cl : Clause =
  Residence_V (Just Adv (mkAdv in_Prep co)) (Just NP p.name) live_V
    in mkCl cl.np cl.vp
\end{verbatim}
\endgroup

For $\mathsf{AWant}$, neither the RGL-based nor the current FN-based implementation can be done in the functor because, in Swedish, the verb \emph{vilja} (`to want') evoking $\mathsf{Desiring\_V2\_Act}$ requires the auxiliary verb \emph{ha} (`to have'). This can be seen as a nested auxiliary frame $\mathsf{Possession}$:

\begingroup
\fontsize{9pt}{11pt}\selectfont
\begin{verbatim}
AWant p obj = mkCl p.name (mkV2 (mkV "want")) obj       -- Eng
Desiring_V2_Act (Just NP p.name) (Just NP obj) want_V2
\end{verbatim}
\endgroup

\begingroup
\fontsize{9pt}{11pt}\selectfont
\begin{verbatim}
AWant p obj = mkCl p.name want_VV (mkVP L.have_V2 obj)  -- Swe
Desiring_VV
  (Just VP (Possession_V2 (Nothing NP) (Just NP obj) have_V2).vp)
  (Just NP p.name) want_VV
\end{verbatim}
\endgroup

Assuming that the auxiliary verb can be optionally used also with other Swedish verbs when applying this frame function, the nested frame could be hidden in the Swedish implementation of $\mathsf{Desiring\_V2\_Act}$. This, however, is not the case with $\mathsf{AWantGo}$ which in both languages requires a main nested frame and, thus, can be put in the functor:

\begingroup
\fontsize{9pt}{11pt}\selectfont
\begin{verbatim}
AWantGo p place = mkCl p.name want_VV (mkVP (mkVP go_V) place.to)
\end{verbatim}
\endgroup

\begingroup
\fontsize{9pt}{11pt}\selectfont
\begin{verbatim}
Desiring_VV (Just VP
  (Motion_V_2 (Just Adv place.to) (Nothing Adv) (Nothing NP) go_V).vp)
  (Just NP p.name) want_VV
\end{verbatim}
\endgroup

At first gleam, the new code might look more complex, but it does not specify how the VP is built, and the same uniform code template is used in all cases. The re-implemented version of Phrasebook accepts and generates the same set of sentences as before.

\subsection{Painting Grammar}
\label{ssec:museum}

The Painting grammar is a part of a large scale Natural Language Generation (NLG) grammar developed for the cultural heritage (CH) domain in order to verbalize data about museum objects stored in an RDF-based ontology \cite{DannellsEtAl2012}. A set of RDF triples (subject-predicate-object expressions) forms the input to the application. As an example, a simplified set of triples representing information about the artwork \emph{Le Général Bonaparte} is:

\begingroup
\fontsize{9pt}{11pt}\selectfont
\begin{verbatim}
<LeGeneralBonaparte> <createdBy> <JacquesLouisDavid>
<LeGeneralBonaparte> <hasDimension> <LeGeneralBonaparteDimesion>
<LeGeneralBonaparte> <hasCreationDate> <LeGeneralBonaparteCreationDate>
<LeGeneralBonaparte> <hasCurrentLocation> <MuseeDuLouvre>
\end{verbatim}
\endgroup

This information is combined by the grammar to generate a coherent text. The function in the abstract syntax that combines the triples is the following:

\begingroup
\fontsize{9pt}{11pt}\selectfont
\begin{verbatim}
DPainting : Painting -> Painter -> Year -> Size -> Museum -> Description
\end{verbatim}
\endgroup

Each argument of the function corresponds to a class in the ontology. Below we show how the arguments are linearized in the original concrete syntax for English and how this syntax has been adapted to generate from the FN-based grammar. To adapt the grammar, we first identified the frames that match the target verbs in the linearization rules. Then we matched the core FEs of the identified frames with the verb arguments.

\begingroup
\fontsize{9pt}{11pt}\selectfont
\begin{verbatim}
The original grammar:               Using the FrameNet-based API:
--------------------------------    -------------------------------------
DPainting painting painter          DPainting painting painter
 year size museum =                  year size museum =
let                                 let
 s1 : Text = mkText (mkS             cl1 : Clause =
  pastTense (mkCl painting (mkVP      Create_physical_artwork_V2_Pass
   (mkVP (passiveVP paint_V2)          (Just NP painter.long)
    (mkAdv by8agent_Prep               (Just NP painting)
     painter.long)) year.s))) ;        paint_V2 ;

 s2 : Text = mkText                  cl2 : Clause = Dimension_V
  (mkCl it_NP (mkVP (mkVP             (Just Adv size.s)
   (mkVPSlash measure_V2)             (Just NP it_NP)
   (mkNP (mkN ""))) size.s) ;         measure_V2 ;

 s3 : Text = mkText                  cl3 : Clause = Being_located_V
  (mkCl (mkNP this_Det painting)      (Just Adv museum.s)
   (mkVP (passiveVP display_V2)       (Just NP (mkNP this_Det painting))
    museum.s))                        display_V2
               
in mkText s1 (mkText s2 s3) ;       in mkText (mkText (mkS pastTense
                                     (mkCl cl1.np (mkVP cl1.vp year.s)))  
                                     (mkText (mkCl cl2.np cl2.vp)
                                      (mkText (mkCl cl1.np cl3.vp))) ;
\end{verbatim}
\endgroup

The grammar exploits patterns of frames $\mathsf{Create\_physical\_artwork}$, $\mathsf{Dimension}$ and $\mathsf{Being\_located}$. 
Since the FN-based grammar currently does not cover non-core FEs, the adjunct \emph{Year} is associated with no FE in $\mathsf{Create\_physical\_artwork}$. Instead, it is attached to the corresponding clause in the final linearization rule ($\mathsf{mkText}$) illustrating how non-core FEs can be incorporated.

The Swedish syntax was adapted in a similar way. The only difference in comparison to English and to the original Swedish syntax is the choice of verbs and pronouns.
The descriptions generated by the new version of $\mathsf{DPainting}$ are semantically equivalent to the descriptions produced by the original grammar:

\begin{quote}
Eng: \emph{Le Général Bonapart was painted by Jacques-Louis David in 1510. It measures 81 by 65 cm. This work is displayed at the Musée du Louvre.}\\
Swe: \emph{Le Général Bonapart målades av Jacques-Louis David år 1510. Den mäter 81 gånger 65 cm. Det här verket hänger på Louvren.}
\end{quote}

\section{Evaluation}
\label{sec:evaluation}

We have conducted a simple intrinsic and extrinsic evaluation of the acquired FN-based grammar. For an initial intrinsic evaluation, we count the number of examples in the source corpora that belong to the set of shared frames and that are covered by the set of shared semantico-syntactic valence patterns. Corpus examples are represented by sentence patterns disregarding non-core FEs, word order and prepositions, but including syntactic roles and the grammatical voice. There are 55,837 examples in BFN that belong to the shared set of 423 frames, and 69.4\% of them are covered by the shared valence patterns despite the modest size of SweFN. In SweFN, 2,434 examples belong to the shared set of frames, and 68.9\% of them are covered by the shared patterns. Note that the original sentences are, in general, covered by paraphrasing.

For an initial extrinsic evaluation, we compare the original application grammars with their FN-based counterparts in terms of code complexity. Since we do not modify the abstract syntax of application grammars, the amount of linearization rules remains the same. Therefore we count the number of constructors used to linearize the functions. In the Painting grammar, the number of constructors is considerably reduced from 21 to 13. In the case of Phrasebook, the number is slightly reduced from 10 in English and 11 in Swedish to 8 in both languages.

\section{Related Work}
\label{sec:related}

The main difference between this work and the previous approaches to CNL grammars is that we present an effort to exploit a robust and well established semantic model in the grammar development. Our approach can be compared with the work on multilingual verbalisation of modular ontologies using GF and \emph{lemon}, the Lexicon Model for Ontologies~\cite{DavisEtAl2012}. 
We use additional lexical information about syntactic arguments for building the concrete syntax. 

The grounding of NLG using the frame semantics theory has been addressed in the work on text-to-scene generation~\cite{CoyneEtAl2011} and in the work on text generation for navigational tasks~\cite{RothAndFrank2010}. 
In that research, the content of frames is utilized through alignment between the frame-semantic structure and the domain-semantic representation. Discourse is supported by applying aggregation and pronominalization techniques. In the CH use case, we also show how an application which utilizes the FN-based grammar can become more discourse-oriented; something that is necessary in actual NLG applications and that has been demonstrated for the CH domain in GF before~\cite{Dannells2010}.
In our current approach, the semantic representation of the domain and the linguistic structures of the grammar are based on FN-annotated data. 

As suggested before~\cite{GruzitisAndBarzdins2010}, a FN-like approach can be used to deal with polysemy in CNL texts. Although we consider lexicalisation alternatives and restrictions for LUs and FEs, we do not address the problem of selectional restrictions and word sense disambiguation in general.

\section{Conclusion}
\label{sec:conclusion}

In this paper we demonstrated the advantages of utilizing a FrameNet-based grammar to facilitate the development of multilingual CNL applications. 
We presented an approach to generating semantic grammar library from two FN-annotated corpora. We tested the feasibility of this grammar as a semantic API for developing application grammars in GF. The major advantage is that language-dependent clause-level specifications to a large extent are hidden by the API, making the application grammars more robust and flexible.

\subsubsection{Acknowledgements.}

This research has been supported by the Swedish Research Council under Grant No. 2012-5746 (Reliable Multilingual Digital Communication: Methods and Applications) and by the Centre for Language Technology in Gothenburg.

\bibliography{cnlBib}
\bibliographystyle{splncs}

\end{document}